# Deducing Optimal Classification Algorithm for Heterogeneous Fabric

Omar Alfarisi, Zeyar Aung, and Mohamed Sassi

**Abstract**
For defining the optimal machine learning algorithm, the decision was not easy for which we shall choose. To help future researchers, we describe in this paper the optimal among the best of the algorithms. We built a synthetic data set and performed the supervised machine learning runs for five different algorithms. For heterogeneous rock fabric, we identified Random Forest, among others, to be the appropriate algorithm.

## 1. Introduction

Among the long list of machine learning algorithms, it was unclear where to start and what machine learning algorithm to use. The selection of the machine learning algorithm is vital to have the machine optimally solve a challenge, as different approaches can deliver different results (*1*). In dealing with heterogeneous systems (*2*) we expect a hybrid system as a complex natural environment (*3*). The choice of the machine learning algorithm is critical, because heterogeneous embeds complex interlinks within its features. Our heterogeneous system is Cretaceous carbonate rock fabric (*4-13*) where we target digitally classifying (*14*) it based on its physical and chemical properties using machine learning. The literature suggested that a decision tree-based algorithm performs better (*15*) than an artificial neural network-based algorithm, like Convolutional Neural Network (CNN), that works better with homogeneous systems (*16, 17*). By homogeneous, we mean measurements conducted in a well-controlled lab environment, at a desirable resolution scale. Therefore, to deduce the optimal classification algorithm for heterogeneous fabric, we suggest Multiple Experimental Stages Selection Learning (MESSL) until reaching the winning algorithm. We built a set of synthetic tabulated data (*18*) to experiment algorithms performance. The data consists of four input properties (independent parameters) and one output (dependent) of four labels. We design the input properties ranges to mimic what a domain expert anticipates from measuring these four properties.

## 2. Method
### 2.1. Selecting Machine Learning Algorithms
We have chosen machine learning algorithms that have different characteristics. We select algorithms K-Nearest Neighbour (KNN), Logistic Regression (LR), Naive Bayes (NB) (*15*), Support Vector Machine (SVM), and Random Forest (RF).

### 2.2. Building Heterogeneous Synthetic Data Set
We made synthetic data representing features in a heterogeneous system, representing microporous media of a Cretaceous rock fabric, which is ~110 million years old formation (*19*). We display a sample of the synthetic data set in Figure 1. We created the feature data using a random number generator function (*20*). We generated the target labels data to mimic domain expert (Geoscientist) decision.

2.2.1. FEATURES DATA

1. PixelColor: Pixel Color, this feature means the quantity or value of the pixel. This feature data range is between 0-255.
2. PhiXSectContin: Pore Cross Section (or black color area size); this feature means that the pore morphology does not have an enclosure from every direction, but it has connections with another black color pixel from two places at least. This feature value ranges 0.00-1.00, where 0.00 means that it has enclosure from all sides (surrounded by non-black color from all sides), while 1.00 means the void is open from all sides (or black pixel is surrounded by black pixels from all directions).
3. NeighbColorGrad: Neighboring Pixel Color Gradient. This feature represents the average gradient of the neighboring Pixels. The range of this feature is between 10-90.
4. Betw2Amplify: Between Two Amplifications. This Feature represents the location property in the black color media (porous media) between the two largest connected black areas. The range of this feature is between 0.00-1.00.

|    | PhiXSectContin | PixelColor | NeighbColorGrad | Betw2Amplify | Lable   |
|----|----------------|------------|-----------------|--------------|---------|
| 0  | 0              | 251        | 64              | 0            | Solid   |
| 1  | 1              | 78         | 19              | 1            | thraot  |
| 2  | 0              | 138        | 29              | 0            | NC_Vugs |
| 3  | 0              | 133        | 35              | 1            | NC_Vugs |
| 4  | 1              | 185        | 45              | 0            | Solid   |
| 5  | 1              | 96         | 84              | 0            | Pore    |
| 6  | 0              | 238        | 43              | 1            | Solid   |
| 7  | 0              | 155        | 51              | 1            | Solid   |
| 8  | 0              | 213        | 67              | 1            | Solid   |
| 9  | 1              | 185        | 67              | 1            | thraot  |
| 10 | 0              | 65         | 81              | 0            | NC_Vugs |
| 11 | 0              | 129        | 30              | 0            | NC_Vugs |
| 12 | 1              | 176        | 66              | 0            | Solid   |
| 13 | 0              | 10         | 47              | 0            | NC_Vugs |
| 14 | 0              | 137        | 45              | 1            | NC_Vugs |
| 15 | 0              | 85         | 12              | 0            | NC_Vugs |
| 16 | 0              | 260        | 26              | 1            | Solid   |
| 17 | 1              | 155        | 22              | 0            | Solid   |
| 18 | 1              | 206        | 53              | 0            | Solid   |
| 19 | 1              | 187        | 51              | 1            | thraot  |

*Figure 1.* Sample of the synthetic data set we generated and used for training and testing several machine learning algorithms. The first column shows the sample number. The second, third, and fourth columns show the synthetic features data we generated that mimic a heterogeneous micropore system image. The last column shows the target labels.

2.2.2. TARGET LABELS DATA
1. Solid: The solid matter of the object.
2. Throat: The object where the diameter of the black cross-sectional area is the smallest.
3. Pore: The black area (pore) in the object where no solid exists.
4. NC-Vugs: The black area (pore) is not connected to other black areas in the image (other pores in the object).

## 2.3. Visualizing the Synthetic Data Set

This visualization step provides an intuitive perspective about the system heterogeneity represented in the synthetic data we generated. The data display is shown in Figure 2. We notice that the classes have a significant overlap, which increases the difficulty of the classification task. This visualization also provides a sound quality control stage before using the data for training and testing the algorithms.

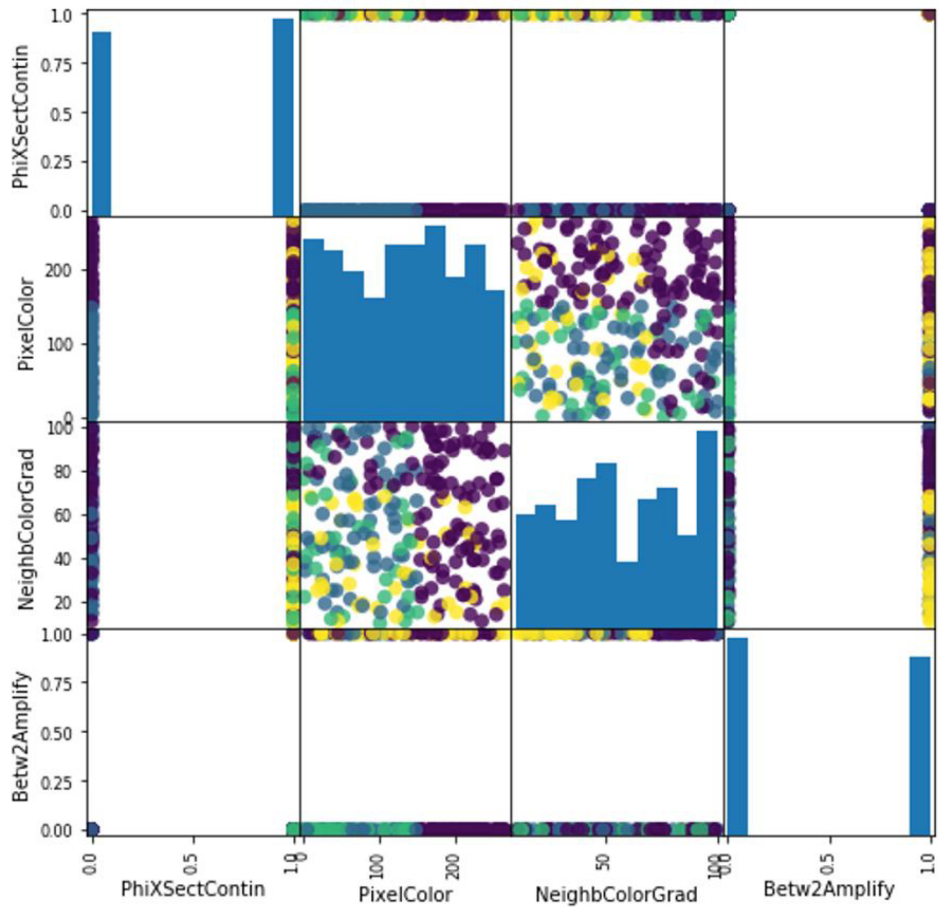

*Figure 2*. The scatter chart and histogram of the synthetic data set. The x-axis and the y-axis show the four features. The colored filled dots show the four target labels. This graph provides an integrated view of the whole data set. The heterogeneity in the data set is observable. This visualization increases the confidence in the data set in the delivery complex system for the machine-learning algorithm to solve. The definitions of the features are in section 2.2.1.

**2.4. Run Synthetic Heterogeneous Data with Five Different Machine Learning Algorithms**

We run the synthetic data to test the capability of the five machine learning algorithms KNN, SMV, LR, NB, and RF, to identify the algorithm that can achieve the highest accuracy for heterogeneous data types. We built the code on Python 3.7 and used the scikit-learn library (*15, 21*). The train and test scores result of the five runs are shown in Figure 3.

**3. Results**

KNN is the loser on our data set. KNN method showed the lowest accuracy among all the other ways with our data set. The K value is set to 1 to deliver the best results; otherwise, the higher the value, the higher the accuracy. Having the classes close to each other's makes it hard for KNN to perform well. The Gaussian Naive Bayes (NB) appeared to be the second-best classifier for our data set. The best performance goes to RF. The setting parameters for RF are the Split (test-size = 0.5), Split (random-state = 3, where random-state is "the seed used by the random number generator,") (*22-24*). The other main parameter that achieved the highest accuracy for Random Forest is n-estimator ("the number of trees in the forest"), where we found that the optimum value, for our data set, is six or higher (n estimators = 6) (*25, 26*).

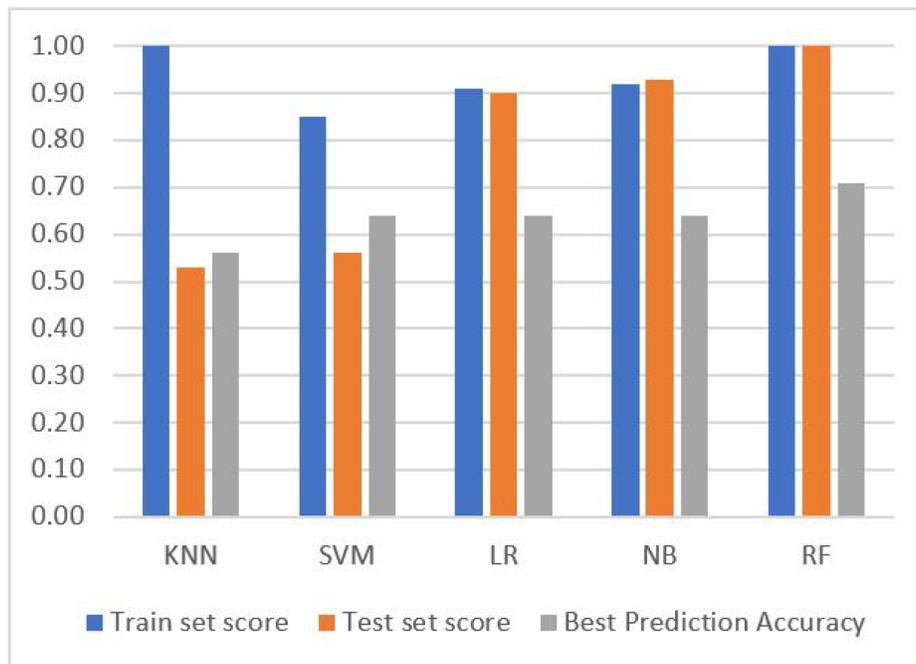

*Figure 3.* Train set score, test set score, and best prediction Accuracy results of five Machine Learning algorithms, using the same synthetic heterogeneous data set we generated, to identify optimal machine learning algorithms in solving systems with heterogeneity, like natural structures and environments. It is important to note that these best prediction accuracy results are collected from the best results achieved out of several runs in which the setting parameters for the machine learning algorithms were optimized.

## 4. Discussion
We noticed that the main controlling factors that improve the predictability are:
> 1. The values of the machine learning setting parameters are one of the main controlling factors, which we adjust to reach the highest possible Accuracy, Recall, Precision, F1, and Test Score. However, we suggest running parameter optimization using Monte-Carlo or another optimizer.
> 2. The Method itself; we noticed that the training and test scores vary from one method to another. Despite the trials of changing the parameters, we can see that the Train and Test Scores differ from one way to another. The best approach appears to be Random Forest, where it showed the best ability to learn and predict.
> 3. The Split ratio between Train and Test data: We can notice that the more the train data ratio, the better the models and the better the prediction. We used test sizes as 0.05 and 0.2 (where the train set as 0.95 and 0.8 respectively) and compared the results. Some methods get better with a minor training set and some better with a more extensive train set.
> 5. Conclusions and Recommendations
> We concluded that Random Forest (RF), a decision tree-based algorithm, is the optimal machine learning algorithm when solving data of the heterogeneous system. Other machine learning algorithms, SVM, LG, and NB, would be the second choice after RF. In comparison, KNN showed the lowest capability in predicting heterogeneity.

Finally, we recommend to the researcher that further improvement of the algorithm's capability would be in using the points below:
> 1. Changing the setting parameters of the different methods further by making the machine learning sequence automated as follows; (1) Train, (2) Test, (3) Predict, (4) Optimize, then go back to the Train and so on until reaching the optimal for that machine learning algorithm.
> 2. Changing the number of data samples by optimizing each method to have the best number of samples that deliver better results. This step can be included in the optimization phase.
> 3. Changing the number of features or improving the features by performing more feature engineering to have better selections. This expert-based optimization can be an additional step after all parameter optimizations.
> 4. Developing hybrid methods that combine human experts' analysis and machine learning in an iterative approach. This can be achieved by following the sequence of (1) Refine Features, (2) Train, (3) Test, (4) Predict, (5) Optimize, then go back to the Train-Test-Predict-Optimize and so on until reaching the optimal for that machine learning algorithm, then go back to Refine Features and so on until the optimal is reached for the machine learning algorithm.

**Affiliation**
Omar Alfarisi (ADNOC Offshore).
Zeyar Aung (Khalifa University).
Mohamed Sassi (Khalifa University).

**Acknowledgement**

The authors appreciate all the support received from ADNOC, ADNOC Offshore, and Khalifa University of Science and Technology in providing the data, enabling laboratory utilization, and other related administrative support. We would like to thank Mr. Yasser Al-Mazrouei, Mr. Ahmed Al-Suwaidi, Mr. Ahmed Al-Hendi, Mr. Mohamed Alghaferi, Mr. Saoud Almehairbi, Mr. Ahmed Al-Riyami, Mr. Andreas Scheed, Mr. Salem Al-Zaabi, and Mr. Mohamed Abdelsalam for their unwavering support, encouragement, and motivation.